\documentclass[11pt,a4paper]{article}
\usepackage{times,latexsym}
\usepackage{url}
\usepackage[T1]{fontenc}

\usepackage[hyperref]{eacl2021}

\usepackage{array}
\usepackage{arydshln}
\usepackage{amsfonts}
\usepackage{amsmath}

\usepackage{bbm}
\usepackage{boldline}
\usepackage{bigstrut}
\usepackage{blindtext}
\usepackage{booktabs, siunitx}
\usepackage[labelfont=bf, format=plain, justification=justified, singlelinecheck=false]{caption}
\usepackage{color}
\usepackage{cprotect}
\usepackage{tikz-dependency, rotating}
\usepackage{ctable}
\usepackage{dirtytalk}
\usepackage{enumitem}
\usepackage[export]{adjustbox}
\usepackage{float}
\usepackage{graphicx}
\usepackage{hyperref}
\usepackage{latexsym}
\usepackage{mathrsfs}
\usepackage{microtype}
\usepackage{moresize}
\usepackage{multicol}
\usepackage{multirow}
\usepackage{nccmath}
\usepackage{nicefrac}
\usepackage{pifont}
\usepackage{placeins}
\usepackage{soul}
\usepackage{subcaption}
\usepackage{tabularx}
\usepackage{times}
\usepackage[utf8]{inputenc}
\usepackage{url}
\usepackage{verbatim}
\usepackage{wrapfig, lipsum}
\usepackage{textcomp}
\usepackage{enumitem}

\newcommand{\fone}{F$_1$}
\newcommand{\ph}[1]{\phantom{#1}}
\newcommand{\cut}[1]{}
\newcommand{\xhdr}[1]{\noindent{\bfseries #1.}}

\aclfinalcopy 
\def\aclpaperid{1021} 

\title{Do Syntax Trees Help Pre-trained Transformers Extract Information?}

\author{Devendra Singh Sachan$^{1,2}$, Yuhao Zhang$^{3}$, Peng Qi$^{3}$, William Hamilton$^{1 ,2}$ \\
$^{1}$Mila - Quebec AI Institute\\
$^{2}$School of Computer Science, McGill University\\
$^{3}$Stanford University\\
{\tt sachande@mila.quebec, wlh@cs.mcgill.ca}\\
{\tt \{yuhaozhang, pengqi\}@stanford.edu}
}

\date{}

\begin{document}
\maketitle


\begin{abstract}
Much recent work suggests that incorporating syntax information from dependency trees can improve task-specific transformer models.
However, the effect of incorporating dependency tree information into \emph{pre-trained} transformer models (e.g., BERT) remains unclear, especially given recent studies highlighting how these models implicitly encode syntax.
In this work, we systematically study the utility of  incorporating dependency trees into pre-trained transformers on three representative information extraction tasks: semantic role labeling (SRL), named entity recognition, and relation extraction.

We propose and investigate two distinct strategies for incorporating dependency structure: a {\em late fusion} approach, which applies a graph neural network on the output of a transformer, and a {\em joint fusion} approach, which infuses syntax structure into the transformer attention layers.
These strategies are representative of prior work, but we introduce additional model design elements that are necessary for obtaining improved performance. 
Our empirical analysis demonstrates that these syntax-infused transformers obtain state-of-the-art results on SRL and relation extraction tasks.
However, our analysis also reveals a critical shortcoming of these models: we find that their performance gains are highly contingent on the availability of human-annotated dependency parses, which raises important questions regarding the viability of syntax-augmented transformers in real-world applications.\footnote{Our code is available at: \url{https://github.com/DevSinghSachan/syntax-augmented-bert}}
\end{abstract}


\section{Introduction} \label{sec:introduction}

Dependency trees---a form of syntactic representation that encodes an asymmetric syntactic relation between words in a sentence, such as \emph{subject} or \emph{adverbial modifier}---have proven very useful in various NLP tasks. 
For instance, features defined in terms of the shortest path between entities in a dependency tree were used in relation extraction (RE)~\cite{fundel2006relex,bjorne2009extracting}, parse structure has improved named entity recognition (NER)~\cite{jie2017efficient}, and joint parsing was shown to benefit semantic role labeling (SRL)~\cite{pradhan-etal-2005-semantic-role} systems. More recently, dependency trees have also led to meaningful performance improvements when incorporated into neural network models for these tasks. Popular encoders to include dependency tree into neural models include graph neural networks (GNNs) for SRL~\cite{marcheggiani-titov-2017-encoding} and  RE~\cite{zhang2018graph}, and biaffine attention in transformers for SRL~\cite{strubell2018srl}.

In parallel, there has been a renewed interest in investigating self-supervised learning approaches to pre-training neural models for NLP, with recent successes including ELMo~\cite{peters2018elmo}, GPT~\cite{radford2018improving}, and BERT~\cite{devlin2019bert}. Of late, the BERT model based on pre-training of a large transformer model~\cite{vaswani2017attn} to encode bidirectional context has emerged as a dominant paradigm, thanks to its improved modeling capacity which has led to state-of-the-art results in many NLP tasks.

BERT's success has also attracted attention to what linguistic information its internal representations capture. For example,~\citet{tenney-etal-2019-bert}  attribute different linguistic information to different BERT layers; \citet{clark-etal-2019-bert} analyze BERT's attention heads to find syntactic dependencies; \citet{hewitt2019structural} show evidence that BERT's hidden representation embeds syntactic trees. 
However, it remains unclear if this linguistic information helps BERT in downstream tasks during finetuning or not. 
Further, it is not evident if external syntactic information can further improve BERT's performance on downstream tasks.

In this paper, we investigate the extent to which pre-trained transformers can benefit from integrating external dependency tree information. We perform the first systematic investigation of how dependency trees can be incorporated into pre-trained transformer models, focusing on three representative information extraction tasks where dependency trees have been shown to be particularly useful for neural models: semantic role labeling (SRL), named entity recognition (NER), and relation extraction (RE).

We propose two representative approaches to integrate dependency trees into pre-trained transformers (i.e., BERT) using syntax-based graph neural networks (syntax-GNNs). 
The first approach involves a sequential assembly of a transformer and a syntax-GNN, which we call \emph{Late Fusion}, while the second approach interleaves syntax-GNN embeddings within transformer layers, termed \emph{Joint Fusion}. 
These approaches are inspired by recent work that combines transformers with external input, but we introduce design elements such as the alignment between dependency tree and BERT wordpieces that lead to obtaining strong performance.
Comprehensive experiments using these approaches reveal several important insights:

\begin{itemize}[topsep=2pt,itemsep=2pt,parsep=2pt, leftmargin=*]
    \item Both our syntax-augmented BERT models achieve a new state-of-the-art on the CoNLL-2005 and CoNLL-2012 SRL benchmarks when the gold trees are used, with the best variant outperforming a fine-tuned BERT model by over 3 F$_1$ points on both datasets. The Late Fusion approach also provides performance improvements on the TACRED relation extraction dataset.

    \item These performance gains are consistent across different pre-trained transformer approaches of different sizes (i.e.\ BERT\textsubscript{BASE/LARGE} and RoBERTa\textsubscript{BASE/LARGE}).

    \item  The Joint Fusion approach that interleaves GNNs with BERT achieves higher performance improvements on SRL, but it is also less stable and more prone to errors when using noisy dependency tree inputs such as for the RE task, where Late Fusion performs much better, suggesting complementary strengths from both approaches.

    \item In the SRL task, the performance gains of both approaches are highly contingent on the availability of human-annotated parses for both training and inference, without which the performance gains are either marginal or non-existent.
    In the NER task, even the gold trees don't show performance improvements.
\end{itemize}

Although our work does obtain new state-of-the-art results on SRL tasks by introducing dependency tree information from syntax-GNNs into BERT, however, our most important result is somewhat negative and cautionary: the performance gains are only substantial when human-annotated parses are available. 
Indeed, we find that even high-quality automated parses generated by domain-specific parsers do not suffice, and we are only able to achieve meaningful gains with human-annotated parses. 
This is a critical finding for future work---especially for SRL---as researchers routinely develop models with human-annotated parses, with the implicit expectation that models will generalize to high-quality automated parses. 

Finally, our analysis provides indirect evidence that pre-trained transformers do incorporate sufficient syntactic information to achieve strong performance on downstream tasks.
While human-annotated parses can still help greatly, with our proposed models it appears that the knowledge in automatically extracted syntax trees is largely redundant with the implicit syntactic knowledge learned by pre-trained models such as BERT.


\section{Models} \label{sec:methods}
In this section, we will first briefly review the transformer encoder, then describe the graph neural network (GNN) that learns syntax representations using dependency tree input, which we term the syntax-GNN. Next, we will describe our syntax-augmented BERT models that incorporate such representations learned from the GNN.

\subsection{Transformer Encoder}
The transformer encoder~\cite{vaswani2017attn} consists of three core modules in sequence: embedding layer, multiple encoder layers, and a task-specific output layer. The core elements in these modules are different sets of learnable weight matrices that perform linear transformations. The embedding layer consists of wordpiece embeddings, positional embeddings, and segment embeddings~\cite{devlin2019bert}. After embedding lookup, these three embeddings are added to obtain token embeddings for an input sentence. The encoder layers then transform the input token embeddings to hidden state representations. The encoder layer consists of two sublayers: multi-head dot-product self-attention and feed-forward network, which will be covered in the following section. Finally, the output layer is task-specific and consists of one layer feed-forward network.

\subsection{Syntax-GNN: Graph Neural Network over a Dependency Tree}
\begin{figure}[t]
\centering
\makebox[\linewidth][c]{\includegraphics[max width=1\linewidth, scale=1.0]{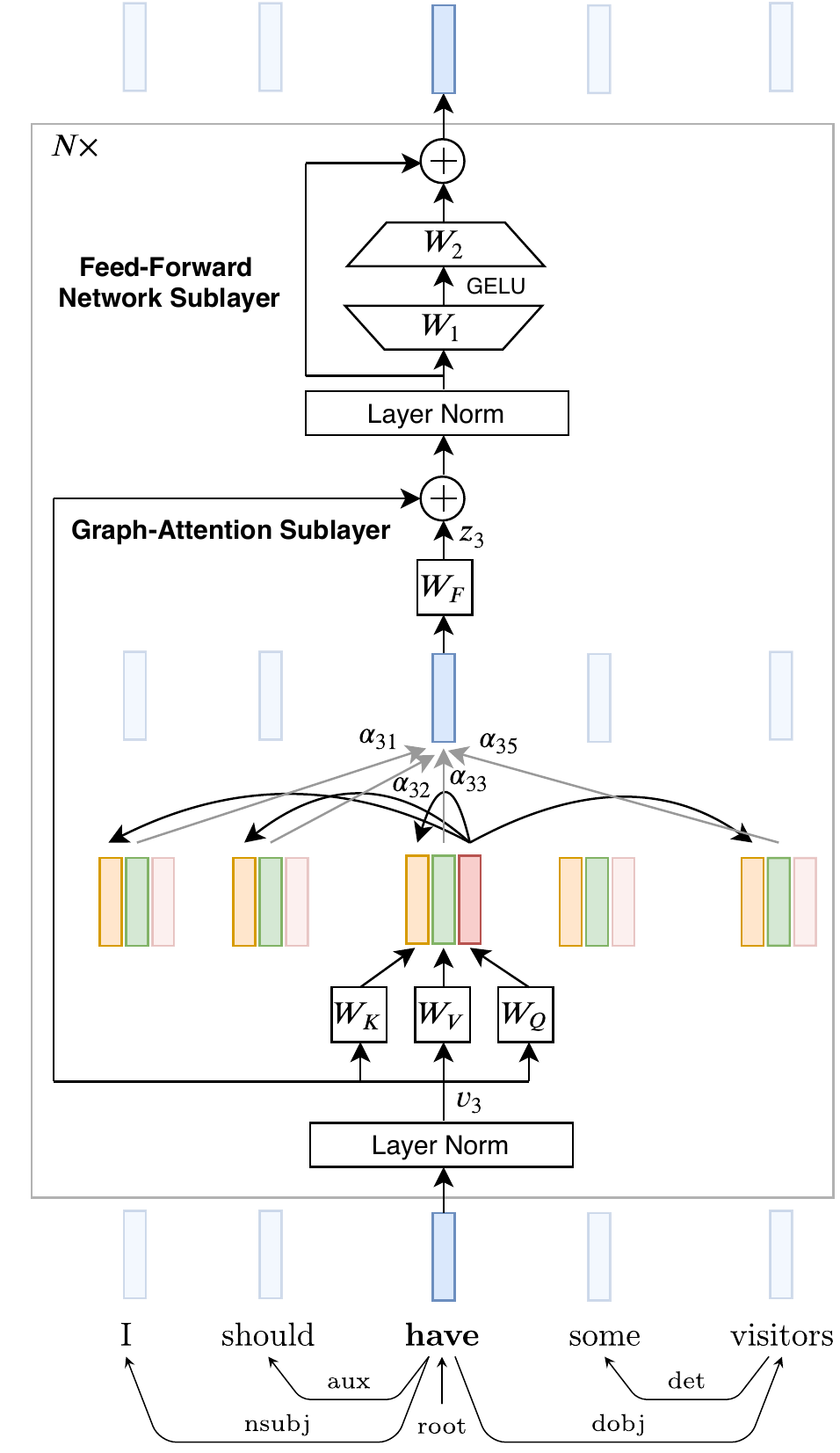}}
\vspace*{-1.7em}
\caption{Block diagram illustrating syntax-GNN applied over a sentence's dependency tree. In the example shown, for the word ``\emph{have}'', the graph-attention sublayer aggregates representations from its three adjacent nodes in the dependency graph.}
\label{fig:syntax-GNN}
\end{figure}

A dependency tree can be considered as a multi-attribute directed graph where the nodes represent words and the edges represent the dependency relation between the head and dependent words. To learn useful syntax representations from the dependency tree structure, we apply graph neural networks (GNNs)~\cite{hamilton2017representation,battaglia2018graph} and henceforth call our model \emph{syntax-GNN}. Our syntax-GNN encoder as shown in Figure~\ref{fig:syntax-GNN} is a variation of the transformer encoder where the self-attention sublayer is replaced by \emph{graph attention}~\cite{velivckovic2017graph}. Self-attention can also be considered as a special case of graph-attention where each word is connected to all the other words in the sentence.

Let $V = \{v_i \in \mathbb{R}^d\}_{i=1:\mathcal{N}^{\textit{v}}}$ denote the input node embeddings and $E = \{(e_k, i, j)_{k=1:\mathcal{N}^{\textit{e}}}\}$ denote the edges in the dependency tree, where the edge $e_k$ is incident on nodes $i$ and $j$. Each layer in our syntax-GNN encoder consists of two sublayers: graph attention and feed-forward network.

First, interaction scores ($s_{ij}$) are computed for all the edges by performing dot-product on the adjacent linearly transformed nodes embeddings
\begin{align} \label{eq:attn_score}
s_{ij} &= (v_i\boldsymbol{W_{\textit{Q}}}) (v_j\boldsymbol{W_{\textit{K}}})^{\top}.
\end{align}
The terms $v_i\boldsymbol{W_{\textit{Q}}}$ and $v_i\boldsymbol{W_{\textit{K}}}$ are also known as \emph{query} and \emph{key} vectors respectively.
Next, an attention score ($\alpha_{ij}$) is computed for each node by applying softmax over the interaction scores from all its connecting edges:
\begin{align}
\alpha_{ij} &= \frac{\exp(s_{ij})}{{\sum_{k\in \mathcal{N}_i}\exp{(s_{ik})}}},
\end{align}
where $\mathcal{N}_\textit{i}$ refers to the set of nodes connected to $i^{\textit{th}}$ node. The graph attention output ($z_i$) is computed by the aggregation of attention scores followed by a linear transformation:
\begin{align}
z_i &= (\sum_{j\in \mathcal{N}_i}\alpha_{ij}(v_j \boldsymbol{W_{\textit{V}}}))\boldsymbol{W_{\textit{F}}}.
\end{align}

The term $v_j \boldsymbol{W_{\textit{V}}}$ is also referred to as \emph{value} vector. Subsequently, the message ($z_i$) is passed to the second sublayer that consists of two layer fully connected feed-forward network with $\mathrm{GELU}$ activation~\cite{hendrycks2016gelu}.
\begin{align}
\mathrm{FFN}(z_i) = \mathrm{GELU}(z_i \boldsymbol{W_1} + b_1) \boldsymbol{W_2} + b_2\ .
\end{align}
The $\mathrm{FFN}$ sublayer outputs are given as input to the next layer. In the above equations $\boldsymbol{W_{\textit{K}}}$, $\boldsymbol{W_{\textit{V}}}$, $\boldsymbol{W_{\textit{Q}}}$, $\boldsymbol{W_{\textit{F}}}$, $\boldsymbol{W_1}$, $\boldsymbol{W_2}$ are trainable weight matrices and $b_1$, $b_2$ are bias parameters. Additionally, layer normalization~\cite{ba2016layer} is applied to the input and residual connections~\cite{he2016deep} are added to the output of each sublayer.

\subsubsection{Dependency Tree over Wordpieces}
As BERT models take as input subword units (also known as wordpieces) instead of linguistic tokens, this also necessitates to extend the definition of a dependency tree to include wordpieces. For this, we introduce additional edges in the original dependency tree by defining new edges from the first subword (head word) of a token to the remaining subwords (tail words) of the same token.

\subsection{Syntax-Augmented BERT}
\label{sec:sf-bert}

\begin{figure}[!t]
\captionsetup[subfigure]{justification=centering}
\centering
\begin{minipage}{0.4\linewidth}
\makebox[\linewidth][l]{\includegraphics[max width=1\linewidth, scale=1.0]{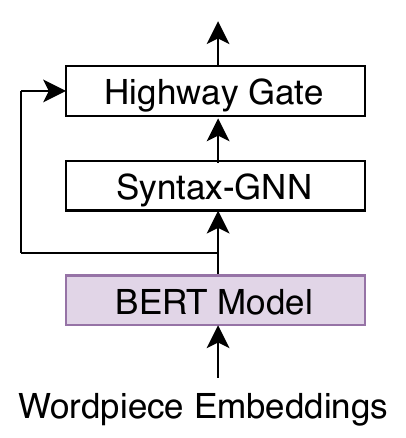}}
\subcaption{Late Fusion}\label{fig:late_fusion}
\end{minipage}%
\begin{minipage}{0.6\linewidth}
\makebox[\linewidth][r]{\includegraphics[max width=1\linewidth, scale=0.9]{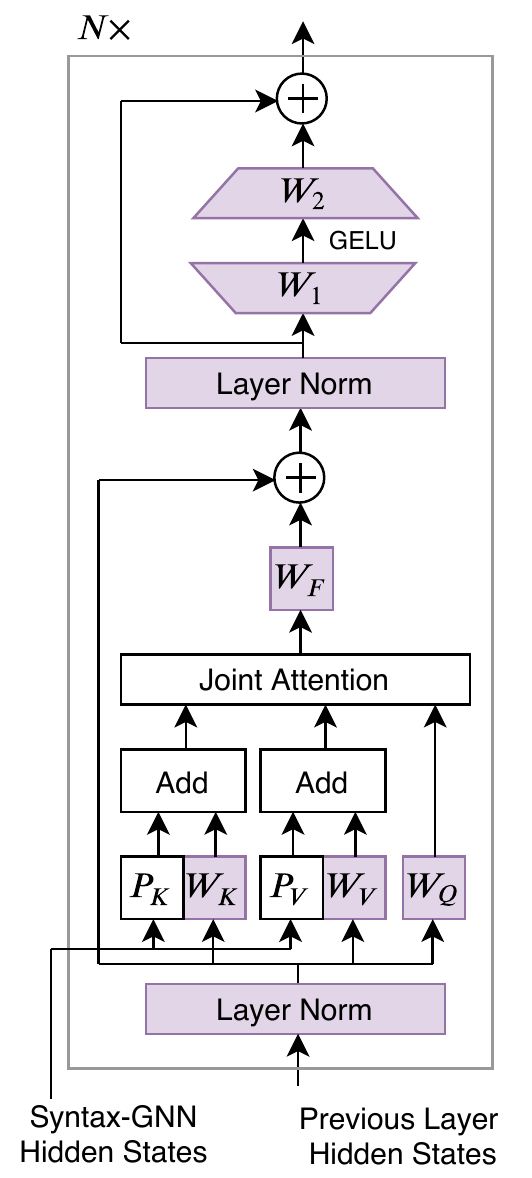}}
  \subcaption{Joint Fusion}\label{fig:joint_fusion}
\end{minipage}
\vspace*{-0.8em}
\caption{Block diagrams illustrating our proposed syntax-augmented BERT models. Weights shown in color are pre-trained while those not colored are either non-parameterized operations or have randomly initialized weights. The inputs to each of these models are wordpiece embeddings while their output goes to task-specific output layers. In subfigure~\ref{fig:joint_fusion}, $N\times$ indicates that there are N layers, with each of them being passed the same set of syntax-GNN hidden states.}
\label{fig:augmented_bert}
\end{figure}

In this section, we propose parameter augmentations over the BERT model to best incorporate syntax information from a syntax-GNN. To this end, we introduce two models--- Late Fusion and Joint Fusion. These models represent novel mechanisms---inspired by previous work---through which syntax-GNN features are incorporated at different sublayers of BERT (Figure~\ref{fig:augmented_bert}). We refer to these models as Syntax-Augmented BERT (SA-BERT) models. During the finetuning step, the new parameters in each model are randomly initialized while the existing parameters are initialized from pre-trained BERT.

\paragraph{Late Fusion:} In this model, we feed the BERT contextual representations to the syntax-GNN encoder \textit{i.e.}\ syntax-GNN is stacked over BERT (Figure~\ref{fig:late_fusion}). We also use a Highway Gate~\cite{srivastava2015highway} at the output of the syntax-GNN encoder to adaptively select useful representations for the training task. Concretely, if $v_i$ and $z_i$ are the representations from BERT and syntax-GNN respectively, then the output ($h_i$) after the gating layer is computed as,
\begin{align}
    g_i &= \sigma(\boldsymbol{W_{\textit g}}v_i + b_g) \\
    h_i &= g_i \odot v_i + (1-g_i)\odot z_i,
\end{align}
where $\sigma$ is the sigmoid function $1/\left( 1 + e^{-x} \right)$ and $\boldsymbol{W_{\textit g}}$ is a learnable parameter. Finally, we map the output representations to linguistic space by adding the hidden states of all the wordpieces that map to the same linguistic token respectively.

\paragraph{Joint Fusion:} In this model, syntax representations are incorporated within the self-attention sublayer of BERT. The motivation is to \textit{jointly} attend over both syntax- and BERT representations. First, the syntax-GNN representations are computed from the input token embeddings and its final layer hidden states are passed to BERT. Second, as is shown in Figure~\ref{fig:joint_fusion}, the syntax-GNN hidden states are linearly transformed using weights $\boldsymbol{P_{\textit K}}$, $\boldsymbol{P_{\textit V}}$ to obtain additional \emph{syntax-based} key and value vectors. Third, syntax-based key and value vectors are added with the BERT's self-attention sublayer key and value vectors respectively. Fourth, the query vector in self-attention layer now attends over this set of keys and values, thereby augmenting the model's ability to fuse syntax information. Overall, in this model, we introduce two new set of weights per layer \{$\boldsymbol{P_{\textit K}}$, $\boldsymbol{P_{\textit V}}$\}, which are randomly initialized.


\section{Experimental Setup} \label{sec:experiments}
\subsection{Tasks and Datasets}
For our experiments, we consider information extraction tasks for which dependency trees have been extensively used in the past to improve model performance. Below, we provide a brief description of these tasks and the datasets used and refer the reader to Appendix~\ref{task_modeling_details} for full details.

\paragraph{Semantic Role Labeling (SRL)} In this task, the objective is to assign semantic role labels to text spans in a sentence such that they answer the query: \emph{Who} did \emph{what} to \emph{whom} and \emph{when}?  Specifically, for every target \emph{predicate} (verb) of a sentence, we detect \emph{syntactic constituents} (arguments) and classify them into predefined semantic roles. In our experiments, we study the setting where the predicates are given and the task is to predict the arguments.
We use the CoNLL-2005 SRL corpus~\cite{carreras2005srl} and CoNLL-2012 OntoNotes\footnote{\textit{\url{conll.cemantix.org/2012/data.html}}} dataset, which contains PropBank-style annotations for predicates and their arguments, and also includes POS tags and constituency parses.

\paragraph{Named Entity Recognition (NER)} NER is the task of recognizing entity mentions in text and tagging them to entity categories. We use the OntoNotes 5.0 dataset~\cite{pradhan-etal-2012-conll}, which contains 18 named entity types. 

\paragraph{Relation Extraction (RE)} RE is the task of predicting the relation between the two entity mentions in a sentence. We use the label corrected version of the TACRED dataset~\cite{zhang2017position,alt2020tacred}, which contains 41 relation types as well as a special \emph{no\_relation} class indicating that no relation exists between the two entities.

\subsection{Training Details}
We select \emph{bert-base-cased} to be our reference pre-trained baseline model.\footnote{\emph{bert-base} configuration was preferred due to computational reasons and we found that \emph{bert-cased} provided substantial gains over \emph{bert-uncased} in the NER task.} It consists of 12 layers, 12 attention heads, and 768 model dimensions. In both the variants, the syntax-GNN component consists of 4 layers, while other configurations are kept the same as \emph{bert-base}. Also, for the Joint Fusion method, syntax-GNN hidden states were shared across different layers. It is worth noting that as our objective is to assess if the use of dependency trees can provide performance gains over pre-trained transformer models, it is important to tune the hyperparameters of these baseline models to obtain strong reference scores. Therefore, for each task, during the finetuning step, we tune the hyperparameters of the default \emph{bert-base} model and use the same hyperparameters to train the SA-BERT models. We refer the reader to Appendix~\ref{add_trainining_details} for additional training details.


\section{Results and Analysis} \label{sec:results}
In this section, we present our main empirical analyses and key findings. 

\begin{table}[t]
\small
    \centering
    \begin{tabular}{l c c c}
    \toprule
    \textbf{Test Set} & \textbf{P} & \textbf{R} & \textbf{\fone} \\
    \midrule
    \multicolumn{4}{c}{\emph{Baseline Models} (without dependency parses)} \\
    \midrule
    SA+GloVe$^\dagger$ & 84.17 & 83.28 & 83.72 \\
    SA+ELMo$^\dagger$ & 86.21 & 85.98 & 86.09 \\
    BERT\textsubscript{BASE} & 86.97 & 88.01 & 87.48 \\
    \midrule
    \multicolumn{4}{c}{\emph{Gold Dependency Parses}} \\
    \midrule
    Late Fusion & 89.17 & 91.09 & 90.12 \\
    Joint Fusion & 90.59 & 91.35 & {\bf 90.97} \\
    \bottomrule
    \end{tabular}
    \vspace*{-0.7em}
    \caption{SRL results on the CoNLL-2005 WSJ test set averaged over $5$ independent runs. $\dagger$ marks results from~\citet{strubell2018srl}. }
    \label{table:conll2005srlwsj_test}
\end{table}

\begin{table}[t]
\small
    \centering
    \begin{tabular}{l c c c}
    \toprule
    \textbf{Test Set} & \textbf{P} & \textbf{R} & \textbf{\fone} \\
    \midrule
    \multicolumn{4}{c}{\emph{Baseline Models} (without dependency parses)} \\
    \midrule
    SA+GloVe$^\dagger$ & 82.55 & 80.02 & 81.26 \\
    SA+ELMo$^\dagger$ & 84.39 & 82.21 & 83.28 \\
    Deep-LSTM+ELMo$^\ddagger$ & - & - & 84.60 \\
    Structure-distilled BERT$^*$ & - & - & 86.39 \\
    BERT\textsubscript{BASE} & 85.91 & 87.07 & 86.49 \\
    \midrule
    \multicolumn{4}{c}{\emph{Gold Dependency Parses}} \\
    \midrule
    Late Fusion & 88.06 & 90.32 & 89.18 \\
    Joint Fusion & 89.34 & 90.44 & {\bf 89.89} \\
    \bottomrule
    \end{tabular}
    \vspace*{-0.7em}
    \caption{SRL results on the CoNLL-2012 test set averaged over $5$ independent runs. $\dagger$ marks results from~\citet{strubell2018srl}; $\ddagger$ mark result from~\citet{peters2018elmo}; $*$ mark result from~\citet{kuncoro2020syntactic}.}
    \label{table:conll2012srl_test}
\end{table}

\begin{table}[t]
\small
    \centering
    \setlength{\tabcolsep}{0.4em}
    \begin{tabular}{l c c c}
    \toprule
    \textbf{Test Set} & \textbf{P} & \textbf{R} & \textbf{\fone} \\
    \midrule
    \multicolumn{4}{c}{\emph{Baseline Models} (without dependency parses)} \\
    \midrule
    BiLSTM-CRF+ELMo$^\dagger$ & 88.25 & 89.71 & 88.98 \\
    BERT\textsubscript{BASE} & 88.75 & 89.61 & 89.18 \\
    \midrule
    \multicolumn{4}{c}{\emph{Gold Dependency Parses}} \\
    \midrule
    DGLSTM-CRF+ELMo$^\dagger$ & 89.59 & 90.17 & {\bf 89.88} \\
    Late Fusion & 88.75 & 89.19 & 88.97 \\
    Joint Fusion & 88.58 & 89.31 & 88.94 \\
    \bottomrule
    \end{tabular}
    \vspace*{-0.7em}
    \caption{NER results on the OntoNotes-5.0 test set averaged over $5$ independent runs. $\dagger$ marks results from~\citet{jie2019dependency}.}
    \label{table:ner_test}
\end{table}

\begin{table}[t]
\small
    \centering
    \setlength{\tabcolsep}{0.6em}
    \begin{tabular}{l c c c}
    \toprule
    \textbf{Test Set} & \textbf{P} & \textbf{R} & \textbf{\fone} \\
    \midrule
    \multicolumn{4}{c}{\emph{Baseline Models} (without dependency parses)} \\
    \midrule
    BERT\textsubscript{BASE} & 78.04 & 76.36 & 77.09 \\
    \midrule
    \multicolumn{4}{c}{\emph{Stanford CoreNLP Dependency Parses}} \\
    \midrule
    GCN$^\dagger$ & 74.2\ph{0} & 69.3\ph{0} & 71.7\ph{0} \\
    GCN+BERT\textsubscript{BASE}$^\dagger$ & 74.8\ph{0} & 74.1\ph{0} & 74.5\ph{0} \\
    Late Fusion & 78.55 & 76.29 & {\bf 77.38} \\
    Joint Fusion & 70.22 & 75.12 & 72.52 \\
    \bottomrule
    \end{tabular}
    \vspace*{-0.7em}
    \caption{Relation extraction results on the revised TACRED test set~\cite{alt2020tacred}, averaged over 5 independent runs. $\dagger$ marks results reported by \citet{alt2020tacred}.
    }
    \label{table:re_test}
\end{table}

\subsection{Benchmark Performance}
To recap, our two proposed variants of the Syntax-Augmented BERT models in Section \ref{sec:sf-bert} mainly differ at the position where syntax-GNN outputs are fused with the BERT hidden states.
Following this, we first compare the effectiveness of these variants on all the three tasks, 
comparing against previous state-of-the-art systems such as~\cite{strubell2018srl,jie2019dependency,zhang2018graph}, which are outlined in Appendix~\ref{additional_baselines} due to space limitations. 
For this part we use gold dependency parses to train the models for SRL and NER, and predicted parses for RE, since gold dependency parses are not available for TACRED.

We present our main results for SRL in Table~\ref{table:conll2005srlwsj_test} and Table~\ref{table:conll2012srl_test}, NER in Table~\ref{table:ner_test}, and RE in Table~\ref{table:re_test}.
All these results report average performance over five runs with different random seeds.
First, we note that for all the tasks, our \emph{bert-base} baseline is quite strong and is directly competitive with other state-of-the-art models.

We observe that both the Late Fusion and Joint Fusion variants of our approach yielded the best results in the SRL tasks. Specifically, on CoNLL-2005 and CoNLL-2012 SRL, Joint Fusion improves over \emph{bert-base} by an absolute $3.5$ \fone{} points, while Late Fusion improves over \emph{bert-base} by $2.65$ \fone{} points.
On the RE task, the Late Fusion model improves over \emph{bert-base} by approximately $0.3$ \fone{} points while the Joint Fusion model leads to a drop of $4.5$ \fone{} points in performance (which we suspect is driven by the longer sentence lengths observed in TACRED).
On NER, the SA-BERT models lead to no performance improvements as their scores lies within one standard deviation to that of \emph{bert-base}.

\emph{Overall, we find that syntax information is most useful to the pre-trained transformer models in the SRL task}, especially when intermixing the intermediate representations of BERT with representations from the syntax-GNN. Moreover, when the fusion is done after the final hidden layer of the pre-trained models, apart from providing good gains on SRL, it also provides small gains on RE task. We further note that, as we trained all our syntax-augmented BERT models using the same hyperparameters as that of \emph{bert-base}, it is possible that separate hyperparameter tuning would further improve their performance.

\subsection{Impact of Parsing Quality}
In this part, we study to what extent parsing quality can affect the performance results of the syntax-augmented BERT models.
Specifically, following existing work, we compare the effect of using parse trees from three different sources:
(a) gold syntactic annotations\footnote{We use Stanford \emph{head rules}~\cite{de-marneffe-manning-2008-stanford} implemented in Stanford CoreNLP v4.0.0~\cite{manning-etal-2014-stanford} to convert constituency trees to dependency trees in UDv2 format~\cite{nivre-etal-2020-universal}.};
(b) a dependency parser trained using gold, \emph{in-domain} parses%
\footnote{The difference between settings (a) and (b) is during test time. In (a) gold parses are used for both training and test instances while in (b) gold parses are used for training, while during test time, parses are extracted from a dependency parser which was trained using gold parses.};
and (c) available off-the-shelf NLP toolkits.\footnote{In this setting, the parsers are trained on general datasets such as the Penn Treebank or the English Web Treebank.} 
In previous work, it was shown that using in-domain parsers can provide good improvements on SRL~\cite{strubell2018srl} and NER tasks~\cite{jie2019dependency}, and the performance can be further improved when gold parses were used at test time.
Meanwhile, in many practical settings where gold parses are not readily available, the only option is to use parse trees produced by existing NLP toolkits, as was done by~\citet{zhang2018graph} for RE.
In these cases, since the parsers are trained on a different domain of text, it is unclear if the produced trees, when used with the SA-BERT models, can still lead to performance gains.
Motivated by these observations, we investigate to what extent \emph{gold}, \emph{in-domain}, and \emph{off-the-shelf} parses can improve performance over strong BERT baselines.

\xhdr{Comparing off-the-shelf and gold parses}
We report our findings on the CoNLL-2005 SRL (Table~\ref{table:conll2005srl_parses}), CoNLL-2012 SRL (Table~\ref{table:conll2012srl_parses}), and OntoNotes-5.0 NER (Table~\ref{table:ner_parses}) tasks.
Using gold parses, both the Late Fusion and Joint Fusion models obtain greater than 2.5 \fone{} improvement on SRL compared with \emph{bert-base} while we don't observe significant improvements on NER. 
We further note that as the gold parses are produced by expert human annotators, these results can be considered as the attainable performance ceiling from using parse trees in these models.

We also observe that using off-the-shelf parses from the Stanza toolkit~\cite{qi2020stanza} provides little to no gains in \fone{} scores (see Tables \ref{table:conll2005srl_parses} and \ref{table:ner_parses}). 
This is mainly due to the low in-domain accuracy of the predicted parses.
For example, on the CoNLL-2012 SRL test set the UAS is 84.2\% for the Stanza parser, which is understandable as the parser was trained on the EWT corpus which covers a different domain.

\begin{figure}[t]
\centering
\begin{minipage}{\linewidth}
\makebox[\linewidth][c]{\includegraphics[max width=0.9\linewidth, scale=0.9]{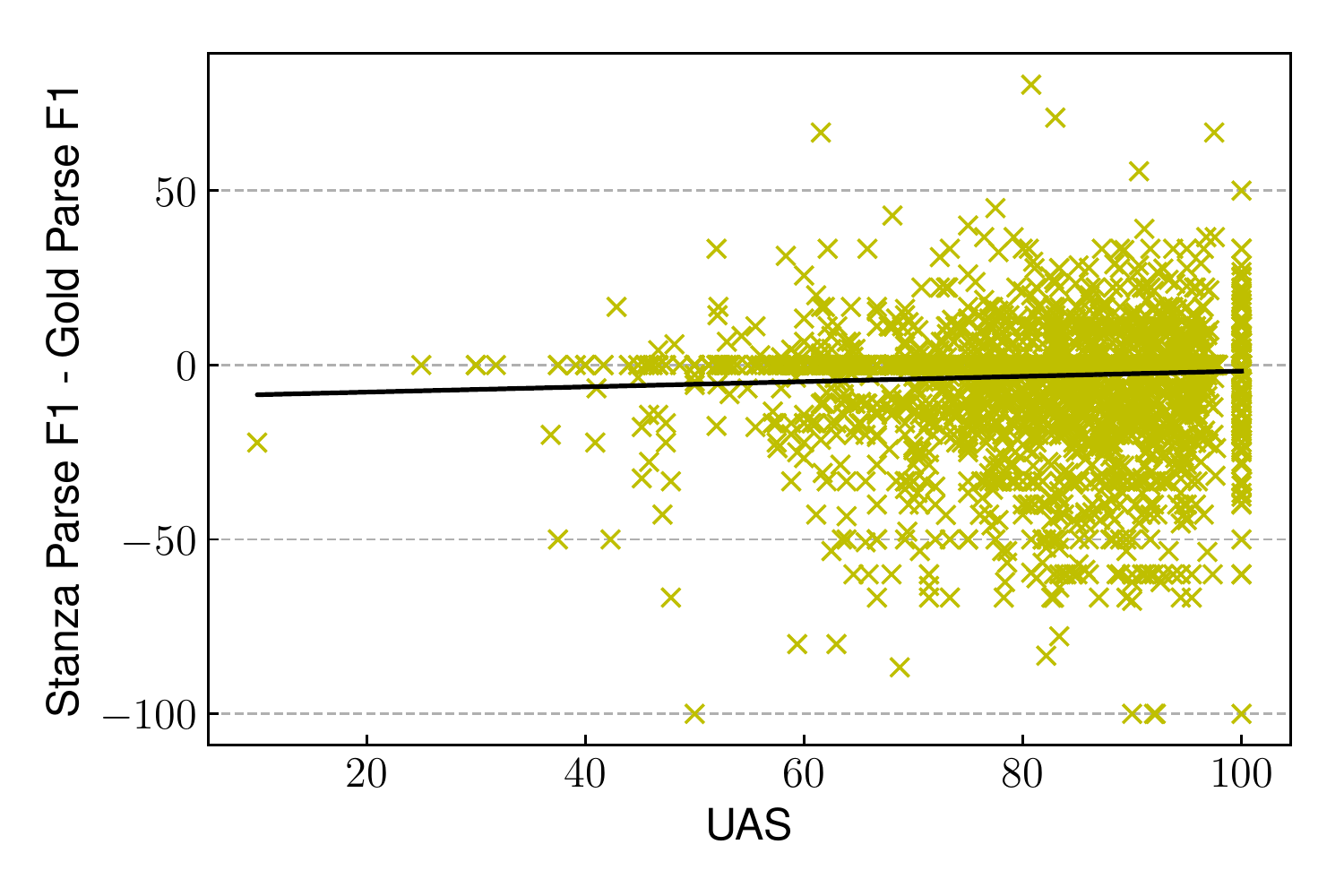}}
\vspace*{-2.em}
\subcaption{When models are trained using Stanza and gold parses, we observe a small positive correlation between \fone{} difference and UAS, suggesting that as UAS of Stanza parse increases, the model makes less errors. The slope of the fitted linear regression model is 0.075 and the intercept is -9.27.
}
\label{fig:F1diffvsUAS}
\end{minipage}
\begin{minipage}{\linewidth}
\centering
\makebox[\linewidth][c]{\includegraphics[max width=0.9\linewidth, scale=0.9]{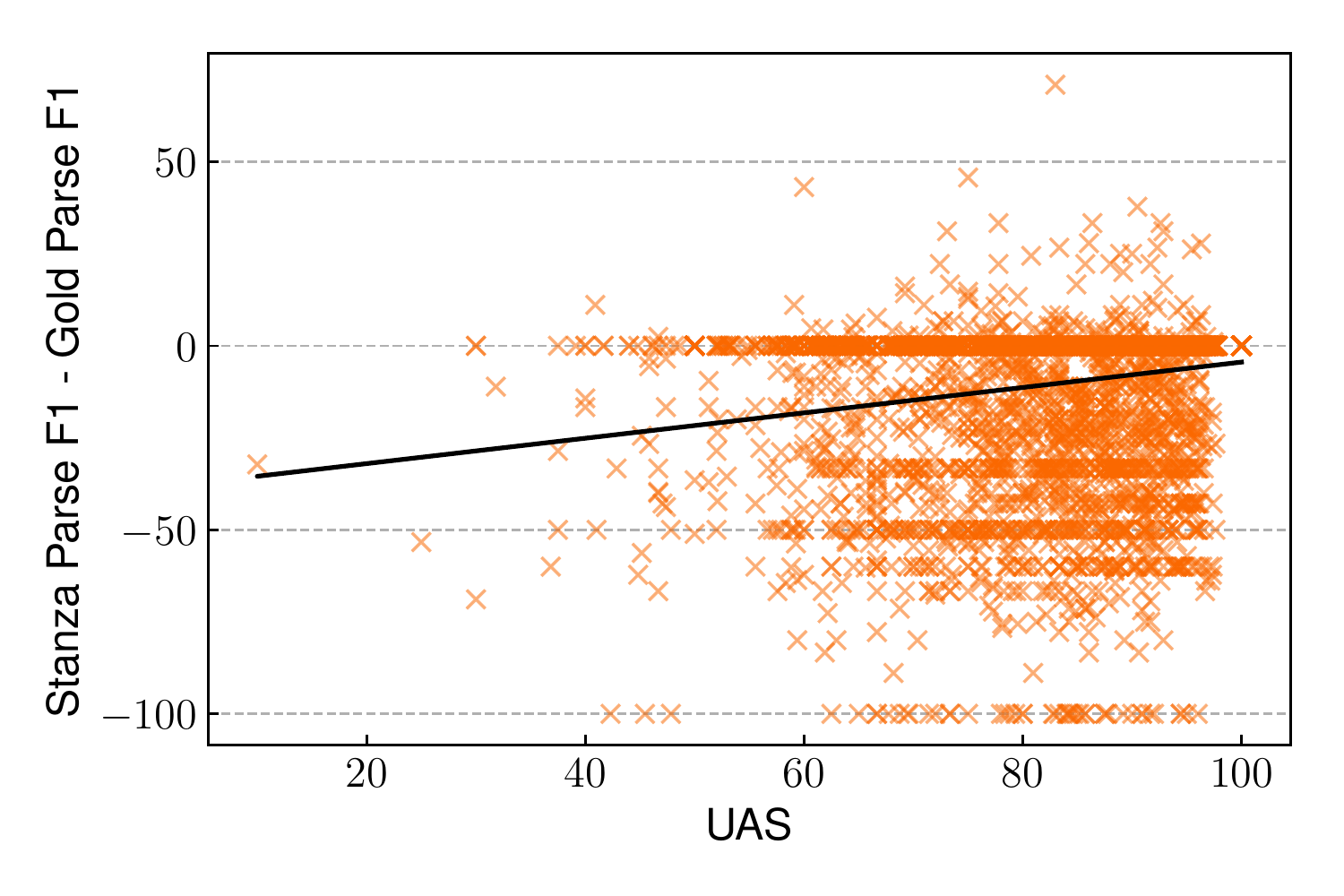}}
\vspace*{-2.em}
\subcaption{Inference is done using Stanza parses on a model trained with gold parses. 
The slope of the fitted linear regression model is 0.345 and the intercept is -38.9.
}
\label{fig:stanza_infer_gold}
\end{minipage}
\caption{Correlation between parse quality and difference in \fone{} scores on CoNLL-2005 SRL WSJ dataset.}
\end{figure}

In a more fine-grained error analysis, we also examined the correlation between parse quality and performance on individual examples on CoNLL-2005 (Figures \ref{fig:F1diffvsUAS} and \ref{fig:stanza_infer_gold}), finding a mild but significant positive correlation between parse quality and relative model performance when training and testing with Stanza parses (Figure \ref{fig:F1diffvsUAS}). 
Interestingly, we found that this correlation between parse quality and validation performance is much stronger when we train a model on gold parses but then evaluate with noisy Stanza parses (Figure \ref{fig:stanza_infer_gold}).
This suggests that the model trained on noisy parses tends to rely less on the noisy dependency tree inputs, while the model trained on gold parses is more sensitive to the external syntactic input.
This correlation is further reinforced by our manual error analysis presented in Appendix~\ref{manual_error_analysis} (Figures~\ref{fig:manual_analysis_srl} and~\ref{fig:manual_error_analysis2}), where we show how the erroneous edges in the Stanza parses can lead to incorrect predictions of the SRL tags.

\begin{table}[t]
\small
    \centering
    \setlength{\tabcolsep}{0.65em}
    \begin{tabular}{l c c c}
    \toprule
    \textbf{Test Set} & \textbf{P} & \textbf{R} & \textbf{\fone} \\
    \midrule
    \multicolumn{4}{c}{\emph{Stanza Dependency Parses} (UAS: 84.20)} \\
    \midrule
    Late Fusion & 86.85 & 88.06 & 87.45 \\
    Joint Fusion & 86.87 & 87.85 & 87.36 \\
    \midrule
    \multicolumn{4}{c}{\emph{In-domain Dependency Parses} (UAS: 92.66)} \\
    \midrule
    LISA+GloVe$^\dagger$ & 85.53 & 84.45 & 84.99 \\
    LISA+ELMo$^\dagger$ & 87.13 & 86.67 & 86.90 \\
    Late Fusion & 86.80 & 87.98 & 87.39 \\
    Joint Fusion & 87.09 & 87.95 & 87.52 \\
    \midrule
    \multicolumn{4}{c}{\emph{Gold Dependency Parses}} \\
    \midrule
    Late Fusion & 89.17 & 91.09 & 90.12 \\
    Joint Fusion & 90.59 & 91.35 & {\bf 90.97} \\
    \bottomrule
    \end{tabular}
    \vspace*{-0.7em}
    \caption{SRL results with different parses on the CoNLL-2005 WSJ test set averaged over $5$ independent runs. $\dagger$ marks results from~\citet{strubell2018srl}.}
    \label{table:conll2005srl_parses}
\end{table}

\begin{table}[t]
\small
    \centering
    \setlength{\tabcolsep}{0.65em}
    \begin{tabular}{l c c c}
    \toprule
    \textbf{Test Set} & \textbf{P} & \textbf{R} & \textbf{\fone} \\
    \midrule
    \multicolumn{4}{c}{\emph{Stanza Dependency Parses} (UAS: 82.73)} \\
    \midrule
    Late Fusion & 85.74 & 87.18 & 86.45 \\
    Joint Fusion & 85.94 & 87.05 & 86.49 \\
    \midrule
    \multicolumn{4}{c}{\emph{In-domain Dependency Parses} (UAS: 93.60)} \\
    \midrule
    Late Fusion & 86.06 & 86.90 & 86.48 \\
    Joint Fusion & 85.75 & 86.92 & 86.33 \\
    \midrule
    \multicolumn{4}{c}{\emph{Gold Dependency Parses}} \\
    \midrule
    Late Fusion & 88.06 & 90.32 & 89.18 \\
    Joint Fusion & 89.34 & 90.44 & {\bf 89.89} \\
    \bottomrule
    \end{tabular}
    \vspace*{-0.7em}
    \caption{SRL results with different parses on the CoNLL-2012 test set.}
    \label{table:conll2012srl_parses}
\end{table}

\begin{table}[t]
\small
    \centering
    \setlength{\tabcolsep}{0.4em}
    \begin{tabular}{l c c c}
    \toprule
    \textbf{Test Set} & \textbf{P} & \textbf{R} & \textbf{\fone} \\
    \midrule
    \multicolumn{4}{c}{\emph{Stanza Dependency Parses} (UAS: 83.91)} \\
    \midrule
    Late Fusion & 88.83 & 89.42 & 89.12 \\
    Joint Fusion & 88.56 & 89.38 & 88.97 \\
    \midrule
    \multicolumn{4}{c}{\emph{In-domain Dependency Parses} (UAS: 96.10)} \\
    \midrule
    DGLSTM-CRF+ELMo$^\dagger$ & -- & -- & 89.64 \\
    \midrule
    \multicolumn{4}{c}{\emph{Gold Dependency Parses}} \\
    \midrule
    DGLSTM-CRF+ELMo$^\dagger$ & 89.59 & 90.17 & {\bf 89.88} \\
    Late Fusion & 88.75 & 89.19 & 88.97 \\
    Joint Fusion & 88.58 & 89.31 & 88.94 \\
    \bottomrule
    \end{tabular}
    \vspace*{-0.7em}
    \caption{NER results with different parses on the OntoNotes-5.0 test set averaged over $5$ independent runs. $\dagger$ marks results from~\citet{jie2019dependency}.}
    \label{table:ner_parses}
\end{table}

\noindent{\bfseries Do in-domain parses help?}
Lastly, for the setting of using in-domain parses, we only evaluate on the SRL task, since on the NER task even using gold parses does not yield substantial gain.
We train a biaffine parser~\cite{dozat2016deep} on the gold parses from the CoNLL-2005 training set and obtain parse trees from it at test time.
We observe that while the obtained parse trees are fairly accurate (with a UAS of 92.6\% on the test set), it leads to marginal or no improvements over \emph{bert-base}. 
This finding is also similar to the results obtained by~\citet{strubell2018srl}, where their LISA+ELMo model only obtains a relatively small improvement over SA+ELMo.
We hypothesize that as the accuracy of the predicted parses further increases, the \fone{} scores would be closer to that from using the gold parses. One possible reason for these marginal gains from using the in-domain parses is that as they are still imperfect, the errors in the parse edges is forcing the model to ignore the syntax information.

Overall, we conclude that \emph{parsing quality has a drastic impact on the performance of the Syntax-Augmented BERT models, with substantial gains only observed when gold parses are used}.

\section{Generalizing to BERT Variants}
Our previous results used the \emph{bert-base} setting, which is a relatively small configuration among pre-trained models.
~\citet{devlin2019bert} also proposed larger model settings (\emph{bert-large}\footnote{24 layers, 16 attention heads, 1024 model dimensions}, \emph{whole-word-masking}\footnote{\textit{\url{https://bit.ly/3l7rbXx}}}) that outperformed \emph{bert-base} in all the benchmark tasks.
More recently, \citet{liu2019roberta} proposed RoBERTa, a better-optimized variant of BERT that demonstrated improved results. 
A research question that naturally arises is: \emph{Is syntactic information equally useful for these more powerful pre-trained transformers, which were pre-trained in a different way than bert-base?} To answer this, we finetune these models---with and without Late Fusion---on the CoNLL-2005 SRL task using gold parses and report their performance in Table~\ref{table:bert_variants}.\footnote{We use the \emph{Late Fusion} model with \emph{gold parses} in this section, as it is computationally more efficient to train than Joint Fusion model.}

As expected, we observe that both \emph{bert-large} and \emph{bert-wwm} models outperform \emph{bert-base}, likely due to the larger model capacity from increased width and more layers. Our Late Fusion model consistently improves the results over the underlying BERT models by about 2.2 \fone. The RoBERTa models achieve improved results compared with the BERT models. And again, our Late Fusion model further improves the RoBERTa results by about 2 \fone{}.
Thus, it is evident that {\em the gains from the Late Fusion model generalize to other widely used pre-trained transformer models}.

\begin{table}[t]
\small
    \centering
    \setlength{\tabcolsep}{0.8em}
    \begin{tabular}{l c c c}
    \toprule
    \textbf{Gold Parses} & \textbf{P} & \textbf{R} & \textbf{\fone} \\
    \midrule
    \multicolumn{4}{c}{\emph{BERT}} \\
    \midrule
    BERT\textsubscript{LARGE} & 88.14 & 88.84 & 88.49 \\
    Late Fusion & 89.86 & 91.57 & {\bf 90.70} \\
    \midrule
    BERT\textsubscript{WWM} & 88.04 & 88.87 & 88.45 \\
    Late Fusion & 89.88 & 91.63 & {\bf 90.75} \\
    \midrule
    \multicolumn{4}{c}{\emph{RoBERTa}} \\
    \midrule
    RoBERTa\textsubscript{LARGE} & 89.14 & 89.90 & 89.47 \\
    Late Fusion & 90.89 & 92.08 & {\bf 91.48} \\
    \bottomrule
    \end{tabular}
    \vspace*{-0.7em}
    \caption{SRL results from using different pre-trained models on the CoNLL-2005 WSJ test set averaged over $5$ independent runs. WWM indicates the whole-wordpiece-masking.}
    \label{table:bert_variants}
\end{table}

\section{Generalizing to Out-of-Domain Data} \label{ood_generalization}
In real-world applications, NLP systems are often used in a domain different from training. And it was previously shown that many NLP systems, such as information extraction systems, suffer from substantial performance degradation when applied to out-of-domain data~\cite{huang-yates-2010-open}.
While it is evident that syntax trees may help models generalize to out-of-domain data \cite{wang2017syntax}, since the inductive biases introduced by these trees are invariant across domains, it is unclear if this hypothesis holds for more recent pre-trained models.
To study this, we run experiments on SRL with the CoNLL-2005 SRL corpus because this is where we have access to both in-domain and out-of-domain test data using the same annotation schema. 
The training set of this corpus contains WSJ articles from the newswire domain and the test set consists of two splits: WSJ articles (in-domain) and Brown corpus\footnote{contains text from 15 genres~\cite{francis79browncorpus}} (out-of-domain). For training, we use both BERT and RoBERTa pre-trained models and leverage gold parses in syntax-GNN models.

From the results in Table~\ref{table:ood_srl}, the utility of syntax-GNN is evident, as we find that the Late Fusion model always improves over its corresponding BERT and RoBERTa baselines by 2-3\% relative \fone, with RoBERTa-large based Late Fusion achieving the best \fone{} on both WSJ and Brown datasets. We also compare the performance across both domains, with the last column showing the relative drop in the \fone{} score between WSJ and Brown datasets. We observe that the performance of all models drops substantially on the Brown set. However, compared with randomly initialized transformer models, where the results can drop by 13\%, both syntax-fused and pre-trained models lead to better generalization as the relative error drop reduces to 6--7\%. \emph{We see that using Late Fusion does not lead to a better out-of-domain generalization, when compared to strong pre-trained transformers without using parse trees.} Lastly, we find that among all pre-trained models, RoBERTa-large and its syntax-fused variant Late Fusion achieves the lowest out-of-domain generalization error.

\begingroup
\begin{table}[t]
\small
    \centering
    \begin{tabular*}{\linewidth}{l c c c c c}
    \toprule
    & \multicolumn{2}{c}{\emph{WSJ Test}} & \multicolumn{2}{c}{\emph{Brown Test}} &  \\
    \midrule
    \textbf{Gold Parses} & \textbf{\fone} & \% $\Delta$ & \textbf{\fone} & \% $\Delta$ & \% $\nabla$ \\
    \midrule
    \multicolumn{6}{c}{\emph{Baseline Models}} \\
    \midrule
    SA+GloVe$^\dagger$ & 84.5 &  & 73.1 &  &  13.5\\
    LISA+GloVe$^\dagger$ & 86.0 & 1.8 & 76.5 & 4.7 & 11.0\\
    \midrule
    \multicolumn{6}{c}{\emph{BERT}} \\
    \midrule
    BERT\textsubscript{BASE} & 87.5 &  & 81.5 & & 6.9\\
    Late Fusion & {\bf 90.1} & 3.0 & {\bf 83.9} & 2.9 & 6.9\\
    \midrule
    BERT\textsubscript{LARGE} & 88.5 &   & 82.5 &  & 6.8\\
    Late Fusion & {\bf 90.8} & 2.6 & {\bf 84.6} & 2.5 & 6.8\\
    \midrule
    \multicolumn{6}{c}{\emph{RoBERTa}} \\
    \midrule
    RoBERTa\textsubscript{LARGE} & 89.5 &  & 84.0 &  & 6.1\\
    Late Fusion & {\bf 91.5} & 2.2 & {\bf 85.5} & 1.8 & 6.6\\
    \bottomrule
    \end{tabular*}
    \vspace*{-0.7em}
    \caption{Out-of-domain SRL results on the CoNLL-2005 WSJ and Brown test sets. $\dagger$ marks results reported in~\citet{strubell2018srl}. \%$\Delta$ denotes the relative gain in \fone{} over pre-trained models when using Late Fusion model, \%$\nabla$ denotes the relative drop in \fone{} when a model trained on WSJ dataset is tested on the Brown dataset.}
    \label{table:ood_srl}
\end{table}
\endgroup


\section{Related Work}  \label{sec:related_work}
Our work is based on finetuning large pre-trained transformer models for NLP tasks, and is closely related to existing work on understanding the syntactic information encoded in them, which we have earlier covered in Section~\ref{sec:introduction}. Here we instead focus on discussing related work that studies incorporating syntax into neural NLP models.

\paragraph{Relation Extraction}
Neural network models have shown performance improvements when shortest dependency path between entities was incorporated in sentence encoders: ~\citet{liu-etal-2015-dependency} apply a combination of recursive neural networks and CNNs;~\citet{miwa2016relation} apply tree-LSTMs for joint entity and relation extraction; and~\citet{zhang2018graph} apply graph convolutional networks (GCN) over LSTM features.

\paragraph{Semantic Role Labeling}
Recently, several approaches have been proposed to incorporate dependency trees within neural SRL models such as learning the embeddings of dependency path between predicate and argument words~\cite{roth-lapata-2016-neural}; combining GCN-based dependency tree representations with LSTM-based word representations~\cite{marcheggiani-titov-2017-encoding}; and linguistically-informed self-attention in one transformer attention head~\cite{strubell2018srl}.
~\citet{kuncoro2020syntactic} directly inject syntax information into BERT pre-training through knowledge distillation, an approach which improves the performance on several NLP tasks including SRL.

\paragraph{Named Entity Recognition}
Moreover, syntax has also been found to be useful for NER as it simplifies modeling interactions between multiple entity mentions in a sentence~\cite{finkel-manning-2009-joint}. To model syntax on OntoNotes-5.0 NER task,~\citet{jie2019dependency} feed the concatenated child token, head token, and relation embeddings to LSTM and then fuse child and head hidden states.


\section{Conclusion} \label{conclusions}
In this work, we explore the utility of incorporating syntax information from dependency trees into pre-trained transformers when applied to information extraction tasks of SRL, NER, and RE. To do so, we compute dependency tree embeddings using a syntax-GNN and propose two models to fuse these embeddings into transformers. Our experiments reveal several important findings: syntax representations are most helpful for SRL task when fused within the pre-trained representations, these performance gains on SRL task are contingent on the quality of the dependency parses. We also notice that these models don't provide any performance improvements on NER. Lastly, for the RE task, syntax representations are most helpful when incorporated on top of pre-trained representations.


\section*{Acknowledgements}
The authors would like to thank Mrinmaya Sachan, Xuezhe Ma, Siva Reddy, and Xavier Carreras for providing us valuable feedback that helped to improve the paper. We would also like to thank the anonymous reviewers for giving us their useful suggestions about this work. This project was supported by academic gift grants from IBM and Microsoft Research, as well as a Canada CIFAR AI Chair held by Prof.\ Hamilton.

\bibliography{main}
\bibliographystyle{acl_natbib_nourl}

\clearpage
\appendix

\section{Experimental Setup}
\subsection{Task-Specific Modeling Details} \label{task_modeling_details}

\paragraph{Semantic Role Labeling (SRL):}
We model SRL as a sequence tagging task using a linear-chain CRF~\cite{lafferty2001crf} as the last layer. During inference, we perform decoding using the Viterbi algorithm~\cite{viterbi}.
To highlight predicate position in the sentence, we use indicator embeddings as input to the model.

\paragraph{Named Entity Recognition (NER):} 
Similar to SRL, we model NER as a sequence tagging task, and use a linear-chain CRF layer over the model's hidden states. Sequence decoding is performed using the Viterbi algorithm.

\paragraph{Relation Extraction (RE):} 
As is common in prior work~\cite{zhang2018graph, miwa2016relation}, the dependency tree is pruned such that the subtree rooted at the lowest common ancestor of entity mentions is given as input to the syntax-GNN. 
Following~\citet{zhang2018graph}, we extract sentence representations by applying a max-pooling operation over the hidden states. We also concatenate the entity representations with sentence representation before the final classification layer.

\subsection{Additional Training Details} \label{add_trainining_details}
During the finetuning step, the new parameters in each model are randomly initialized while the existing parameters are initialized from pre-trained BERT. For regularisation, we apply dropout~\cite{srivastava2014dropout} with $p=0.1$ to attention coefficients and hidden states. For all datasets, we use the canonical training, development, and test splits. We use the Adam optimizer~\cite{kingma2014adam} for finetuning.

We observed that the initial learning rate of 2e-5 with a linear decay worked well for all the tasks. For the model training to converge, we found that $10$ epochs were sufficient for CoNLL-2012 SRL and RE and $20$ epochs were sufficient for CoNLL-2005 SRL and NER. We evaluate the test set performance using the best-performing checkpoint on the development set.

For evaluation, following convention we report the micro-averaged precision, recall, and \fone{} scores in every task. For variance control in all the experiments, we report the mean of the results obtained from five independent runs with different seeds.

\section{Additional Baselines} \label{additional_baselines}
Besides BERT models, we also compare our results to the following previous work, which had obtained good performance gains on incorporating dependency trees with neural models:
\begin{itemize}[topsep=2pt,itemsep=0pt,parsep=2pt, leftmargin=*]
\item For SRL, we include results from the SA (self-attention) and LISA (linguistically-informed self-attention) model by~\citet{strubell2018srl}. In LISA, the attention computation in one attention-head of the transformer is biased to enforce dependent words only attend to their head words. The models were trained using both GloVe~\cite{pennington2014glove} and ELMo embeddings.
\item For NER, we report the results from~\citet{jie2019dependency}, where they concatenate the child token, head token, and relation embeddings as input to an LSTM and then fuse child and head hidden states.
\item For RE, we report the results of the GCN model from~\citet{zhang2018graph} where they apply graph convolutional networks on pruned dependency trees over LSTM states.
\end{itemize}

\section{Manual Error Analysis} \label{manual_error_analysis}
\begin{figure*}
\captionsetup[subfigure]{justification=centering}
\centering
\begin{minipage}{\linewidth}
\input{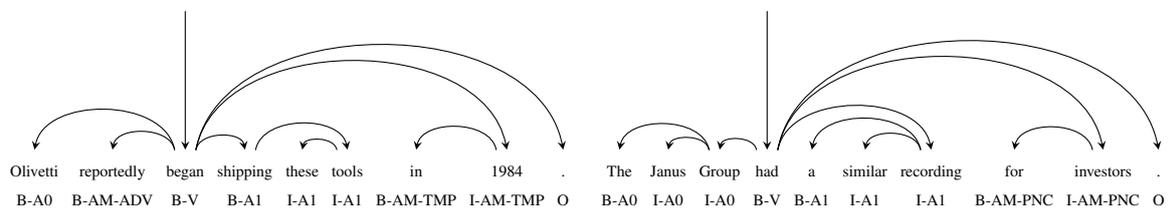}
\vspace*{-0.7em}
\subcaption{Predicted SRL tags using \textbf{Gold parses}} \label{fig:gold}
\end{minipage}
\begin{minipage}{\linewidth}
\input{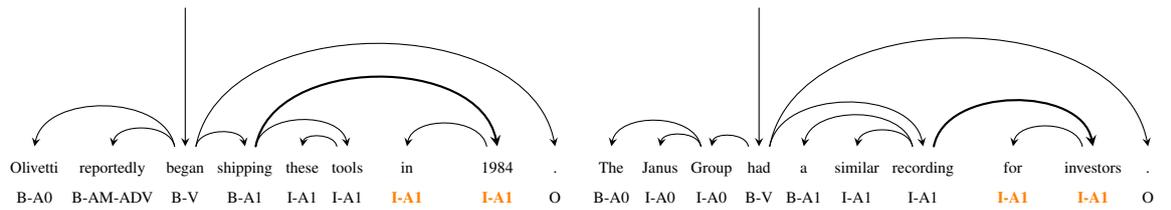}
\vspace*{-0.7em}
\subcaption{Predicted SRL tags using \textbf{Stanza parses}} \label{fig:stanza}
\end{minipage}
\vspace*{-0.5em}
\caption{Examples of sentences with their predicted SRL tags when the Late Fusion model is trained using gold parses (\ref{fig:gold}) and Stanza parses (\ref{fig:stanza}). While the predicted SRL tags using the gold parses are accurate, the erroneous edges in the Stanza parses (highlighted in bold) leads to incorrect SRL tags predictions (highlighted in orange color).}
\label{fig:manual_analysis_srl}
\end{figure*}

\begin{sidewaysfigure*}
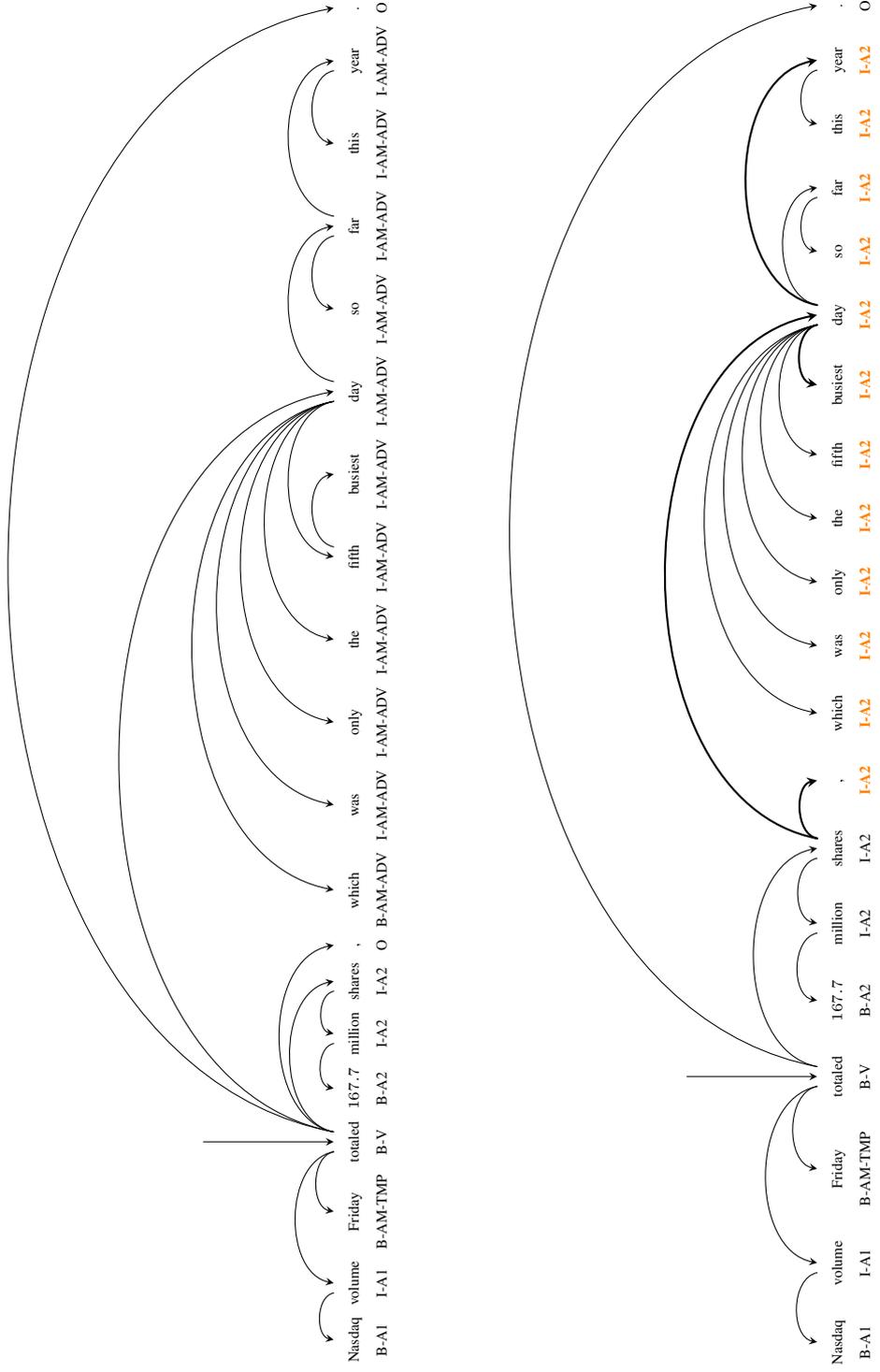

\centering
\begin{dependency}[theme=simple,font=\tiny]
    \begin{deptext}
    Nasdaq \& volume \& Friday \& totaled \& $167.7$ \& million \& shares \& , \& which \& was \& only \& the \& fifth \& busiest \& day \& so \& far \& this \& year \& . \\
    B-A1 \& I-A1 \& B-AM-TMP \& B-V \& B-A2 \& I-A2 \& I-A2 \& O \& B-AM-ADV \& I-AM-ADV \& I-AM-ADV \& I-AM-ADV\& I-AM-ADV \& I-AM-ADV \& I-AM-ADV \& I-AM-ADV \& I-AM-ADV \& I-AM-ADV \& I-AM-ADV \& O \\
    \end{deptext}
    \deproot{4}{}
\depedge{2}{1}{}
\depedge{4}{2}{}
\depedge{4}{3}{}
\depedge{6}{5}{}
\depedge{7}{6}{}
\depedge{4}{7}{}
\depedge{4}{8}{}
\depedge{15}{9}{}
\depedge{15}{10}{}
\depedge{15}{11}{}
\depedge{15}{12}{}
\depedge{15}{13}{}
\depedge{13}{14}{}
\depedge{4}{15}{}
\depedge{17}{16}{}
\depedge{15}{17}{}
\depedge{19}{18}{}
\depedge{17}{19}{}
\depedge{4}{20}{}
\end{dependency}

\begin{dependency}[theme=simple,font=\tiny]
    \begin{deptext}[column sep=0.32cm]
        Nasdaq \& volume \& Friday \& totaled \& $167.7$ \& million \& shares \& , \& which \& was \& only \& the \& fifth \& busiest \& day \& so \& far \& this \& year \& . \\
        B-A1 \& I-A1 \& B-AM-TMP \& B-V \& B-A2 \& I-A2 \& I-A2 \& \textbf{\textcolor{orange}{I-A2}} \& \textbf{\textcolor{orange}{I-A2}} \& \textbf{\textcolor{orange}{I-A2}} \& \textbf{\textcolor{orange}{I-A2}} \& \textbf{\textcolor{orange}{I-A2}} \& \textbf{\textcolor{orange}{I-A2}} \& \textbf{\textcolor{orange}{I-A2}} \& \textbf{\textcolor{orange}{I-A2}} \& \textbf{\textcolor{orange}{I-A2}} \& \textbf{\textcolor{orange}{I-A2}} \& \textbf{\textcolor{orange}{I-A2}} \& \textbf{\textcolor{orange}{I-A2}} \& O \\
    \end{deptext}
    \deproot{4}{}
\depedge{2}{1}{}
\depedge{4}{2}{}
\depedge{4}{3}{}
\depedge{6}{5}{}
\depedge{7}{6}{}
\depedge{4}{7}{}
\depedge[style=thick]{7}{8}{}
\depedge{15}{9}{}
\depedge{15}{10}{}
\depedge{15}{11}{}
\depedge{15}{12}{}
\depedge{15}{13}{}
\depedge[style=thick]{15}{14}{}
\depedge[style=thick]{7}{15}{}
\depedge{17}{16}{}
\depedge{15}{17}{}
\depedge{19}{18}{}
\depedge[style=thick]{15}{19}{}
\depedge{4}{20}{}
\end{dependency}
\caption{Example of a longer sentence with its predicted SRL tags when the Late Fusion model is trained using gold parses (above figure) and Stanza parses (lower figure). The erroneous edges in the Stanza parses are highlighted in bold. While the predicted SRL tags using the gold parses are accurate, the erroneous edges in the Stanza parses leads to a series of incorrect SRL tag predictions (highlighted in orange color).}
\label{fig:manual_error_analysis2}
\end{sidewaysfigure*}

In this section, we present several examples from our manual error analysis of the predictions from the Late Fusion model when it is trained on CoNLL-2005 SRL WSJ dataset using gold and Stanza parses. Specifically, we show how the incorrect edges present in the parse tree can induce wrong SRL tag predictions. In Figure~\ref{fig:manual_analysis_srl}, we observe two examples where the model when trained with gold parses outputs perfect predictions but the when trained with Stanza parses outputs two incorrect SRL tags due to one erroneous edge present in the dependency parse. In Figure~\ref{fig:manual_error_analysis2}, we show an example of a longer sentence where due to the presence of four erroneous edges in the Stanza parse, the model makes a series of incorrect predictions of the SRL tags.

\end{document}